\documentclass[fleqn,10pt]{wlscirep}
\usepackage[utf8]{inputenc}
\usepackage[T1]{fontenc}
\title{Leveraging generative artificial intelligence to simulate student learning behavior}

\author[1,*]{Songlin Xu}
\author[2]{Xinyu Zhang}
\affil[1]{University of California San Diego}
\affil[2]{University of California San Diego}

\affil[*]{soxu@ucsd.edu}


\begin{abstract}
Student simulation presents a transformative approach to enhance learning outcomes, advance educational research, and ultimately shape the future of effective pedagogy. 
We explore the feasibility of using large language models (LLMs), a remarkable achievement in AI, to simulate student learning behaviors. 
Unlike conventional machine learning based prediction, we leverage LLMs to instantiate virtual students with specific demographics and uncover intricate correlations among learning experiences, course materials, understanding levels, and engagement. 
Our objective is not merely to predict learning outcomes but to replicate learning behaviors and patterns of real students. 
We validate this hypothesis through three experiments. 
The first experiment, based on a dataset of N = 145, simulates student learning outcomes from demographic data, revealing parallels with actual students concerning various demographic factors. 
The second experiment (N = 4524) results in increasingly realistic simulated behaviors with more assessment history for virtual students modelling.
The third experiment (N = 27), incorporating prior knowledge and course interactions, indicates a strong link between virtual students' learning behaviors and fine-grained mappings from test questions, course materials, engagement and understanding levels. 
Collectively, these findings deepen our understanding of LLMs and demonstrate its viability for student simulation, empowering more adaptable curricula design to enhance inclusivity and educational effectiveness.
\end{abstract}
\begin{document}

\flushbottom
\maketitle
%
%
\thispagestyle{empty}

In an era defined by rapid technological advancement and an escalating intricacy of global challenges, education stands as the key to unlocking human potential, fostering innovation, and addressing the myriad issues facing our world \cite{trilling200921st,tracey2016educational}. 
However, in the evolution of traditional education, 
it is increasingly clear that education cannot be a one-size-fits-all endeavor\cite{gavsevic2016learning}, and the key to success lies in tailoring the learning experience to the unique needs and characteristics of each student\cite{salend2010creating}. Student simulation \cite{vanlehn1994applications,cherryholmes1966some}, like a digital twin \cite{8901113} in classroom, presents an innovative approach in this regard, allowing us to bridge the gap between education research theory and practice, providing educators with a powerful tool to develop adaptive, personalized curricula \cite{kinsner2021digital} to enhance inclusivity and enrich the pedagogical landscape. 

However, prevailing approaches treat student simulation as a black-box machine learning problem \cite{piech2015deep,beck2000high,hussain2019using,xu2017machine}, overlooking the intricacies of learning process and failing to account for the impact of student learning experiences. 
We posit that student simulation should not be reduced to a mere test score prediction task, as it should encompass vital components of learning behaviors \cite{shelton2017predicting}, including students' demographic information, understanding of course concepts and classroom engagement. Furthermore, given the multifaceted nature of human behaviors, achieving precise predictions of student behaviors becomes exceedingly challenging \cite{cziko1989unpredictability}.

To address these complexities, we propose the development of a digital twin to replicate the nuanced learning behaviors and patterns exhibited by students, as opposed to striving for exact predictions of their behaviors. For instance, the digital twin need not precisely predict the same test scores as real students but should exhibit similar responses when receiving feedback from course instructors, mirroring the behavior of actual students. We examine the potential of utilizing large language models (LLMs) \cite{zhao2023survey}, an achievement in generative artificial intelligence, for simulating student learning behaviors. 
LLMs, advanced AI systems pretrained on extensive datasets, possess the capacity for large-scale natural language understanding and generation, thus enabling the creation of a digital twin to emulate virtual students with specified demographic data, observations, and actions.


We validate our hypothesis through three experiments. 
The first experiment is based on an N = 145 dataset to simulate student learning outcomes by instantiating virtual students with more than 20 kinds of demographic information (such as high school level, living conditions, parents education, etc). The results demonstrate consistent impact of various demographic factors on virtual students' learning outcomes compared with real students. 
In the second experiment (N = 4524), we incorporate assessment history into the simulation, and find increasingly realistic simulated student learning outcomes as more assessment history is used to model virtual students. 
In the third experiment (N = 27), we simulate both learning experiences and outcomes at a fine-grained level by considering students' engagement, prior knowledge, and understanding levels when interacting with course materials. The results indicate that virtual students' learning experiences are linked to their past experiences, and learning outcomes are more realistically simulated when accounting for fine-grained mappings from test questions to corresponding course materials and student understanding levels. In contrast to a coarse-grained approach that aggregates all course materials and understanding levels, this finer-grained analysis offers a more realistic simulation.

Cumulatively, these findings enrich our comprehension of large language models and establish the feasibility and effectiveness of employing AI to simulate student learning behaviors, thus enabling educators to design more engaging, adaptable curricula that promote inclusivity and educational efficacy. Subsequent sections present our primary results and in-depth analysis from the three experiments.

\begin{figure*}
\centering
\includegraphics[width=1\linewidth]{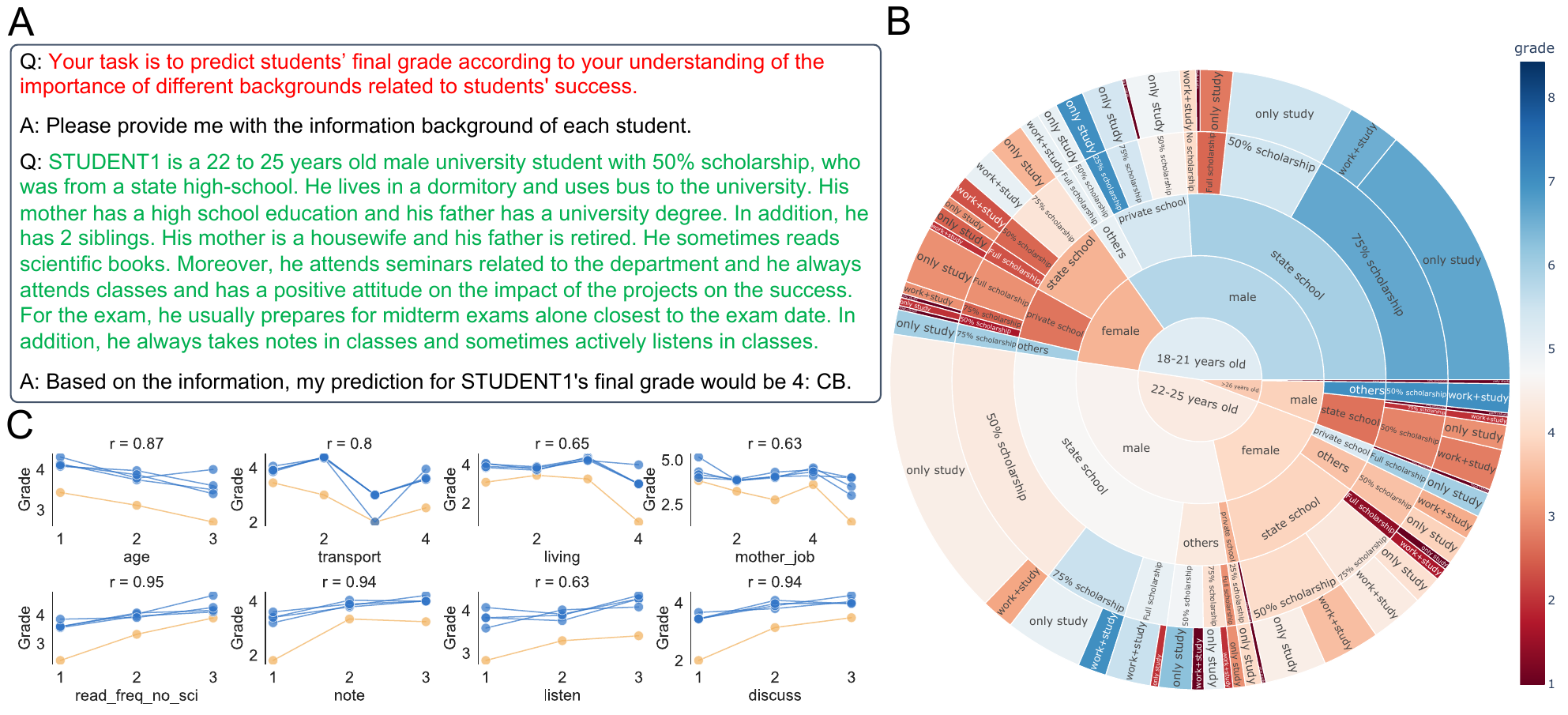}
\caption{(A). Example of input prompts (denoted as Q) and LLMs' responses (denoted as A) in experiment 1. Contents have been truncated. Full examples can be found in Appendix. (B). Example distributions of demographic information in the dataset of experiment 1. (C). Blue and orange dots represent simulation and real human results, respectively. The Pearson coefficient $r$ represents the average correlation in all running times. 
}
\label{f1}
\end{figure*}

\section*{Results}

We employed the OpenAI API \cite{openaicite}, GPT-3.5-Turbo model (referred to as ChatGPT), for conducting our student simulations. The temperature parameter was consistently set at 0 to yield deterministic results, while all other parameters remained at their default settings, unless explicitly specified.

\subsection*{Experiment 1}
\label{sec: R1}

The first experiment aims to examine whether LLMs could capture correlation between student  background and learning outcome. We utilized the dataset from \cite{yilmaz2019student} that includes N = 145 students' demographics (Fig. \ref{f1}(B)) and final grade. A complete list (about 30 kinds of factors) and distributions are depicted in Appendix. We ask LLMs to mimic students with specific demographic backgrounds and  predict the virtual students' final grade (one from 0: Fail, 1: DD, 2: DC, 3: CC, 4: CB, 5: BB, 6: BA, 7: AA). One example prompt and  responses are depicted in Fig. \ref{f1}(A).

Intuitively, it is difficult to directly predict students' final grade without training data. However, as mentioned before, our goal is not to precisely predict students' learning outcomes but to test whether virtual students could exhibit similar patterns as real students. Therefore, instead of directly comparing the simulated final grade accuracy, we validate the correlation between virtual and real students across different demographic factors (Fig. \ref{f1}). Interestingly, we find highly aligned (Pearson correlation: $r$ $>$ 0.7) behaviors in most factors such as age, transportation, living condition, parents jobs, learning habits, etc, as is depicted in Fig. \ref{f1}(C).


To check the robustness of the LLMs simulation, we set temperature parameter to be default value (1 instead of 0) and run the experiment for four times. The simulation results are consistent, as depicted in Fig. \ref{f1}(C). More detailed Pearson coefficients in each demographic factor are depicted in Appendix, where we find significant consistency ($r$ $>$ 0.7) among different study running times in most factors, suggesting the robustness of LLMs' simulation.


These results demonstrate the effectiveness of harnessing LLMs to simulate students' learning outcomes. Specifically, LLMs could capture the correlation between various demographics and students' learning outcomes.
This capacity stems from LLMs' robust in-context learning abilities and extensive knowledge base \cite{dong2022survey}.
Nonetheless, we observe inconsistent correlations between virtual and real students in some demographic factors (see Appendix). Such discrepancies are expected and justifiable given the complexity of human behaviors, making it nearly impossible to replicate precise student behaviors \cite{cziko1989unpredictability}. Additionally, limited training data contributes to this variability. However, this diversity and divergence in simulation outcomes contribute to the development of a more effective digital twin, as determinism alone may not adequately represent the range of student learning behaviors.

\begin{figure*}
\centering
\includegraphics[width=1\linewidth]{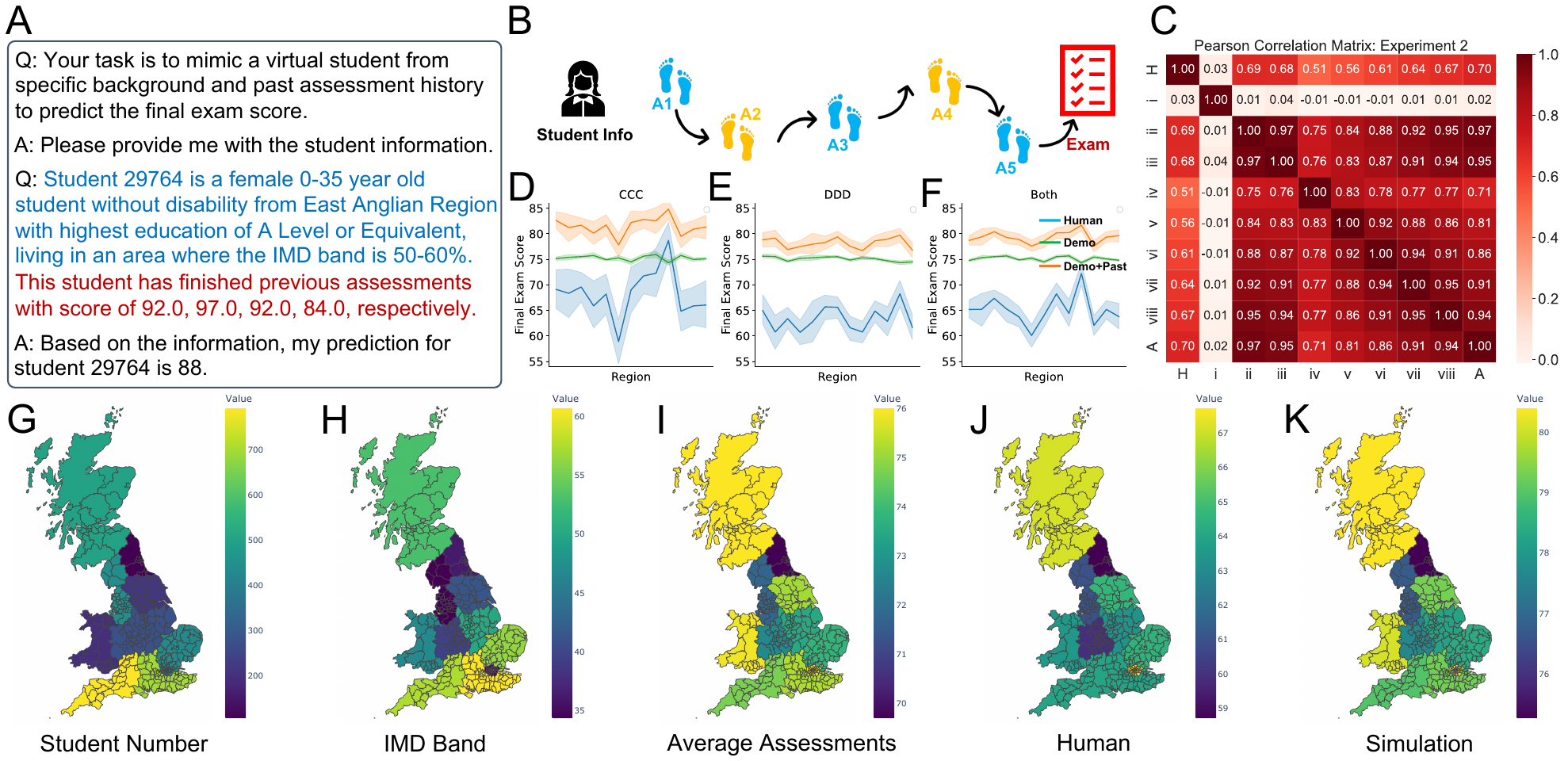}
\caption{(A). Example of input prompts (denoted as Q) and LLMs' responses (denoted as A) in experiment 2. Contents have been truncated. Full examples can be found in Appendix. (B). Illustration of experiment 2 design: virtual students are instantiated with demographics and past assessments history. (C). Pearson correlation matrix in experiment 2 results. i to viii represent simulated final exam scores using the simulation model from Type i to Type viii, respectively. H represents real students' final exam scores and A represents average score of all assessments of real students. (D),(E),(F) represent the distribution of final exam scores in different regions in course id $CCC$ (D), course id $DDD$ (E) and both courses (F), respectively. Blue, green, and orange represent real students, simulated students using only demographics (Type i) and simulated students using both demographics and all past assessments scores (Type iii), respectively. (G),(H),(I),(J),(K) represent the distribution of student number (G), IMD band (H), average assessments score (I), final exam scores of real students (J), and final exam scores of simulated students using Type iii (K) in different regions, respectively.
}
\label{f2}
\end{figure*}

\subsection*{Experiment 2}
\label{sec: R2}

\begin{figure*}
\centering
\includegraphics[width=1\linewidth]{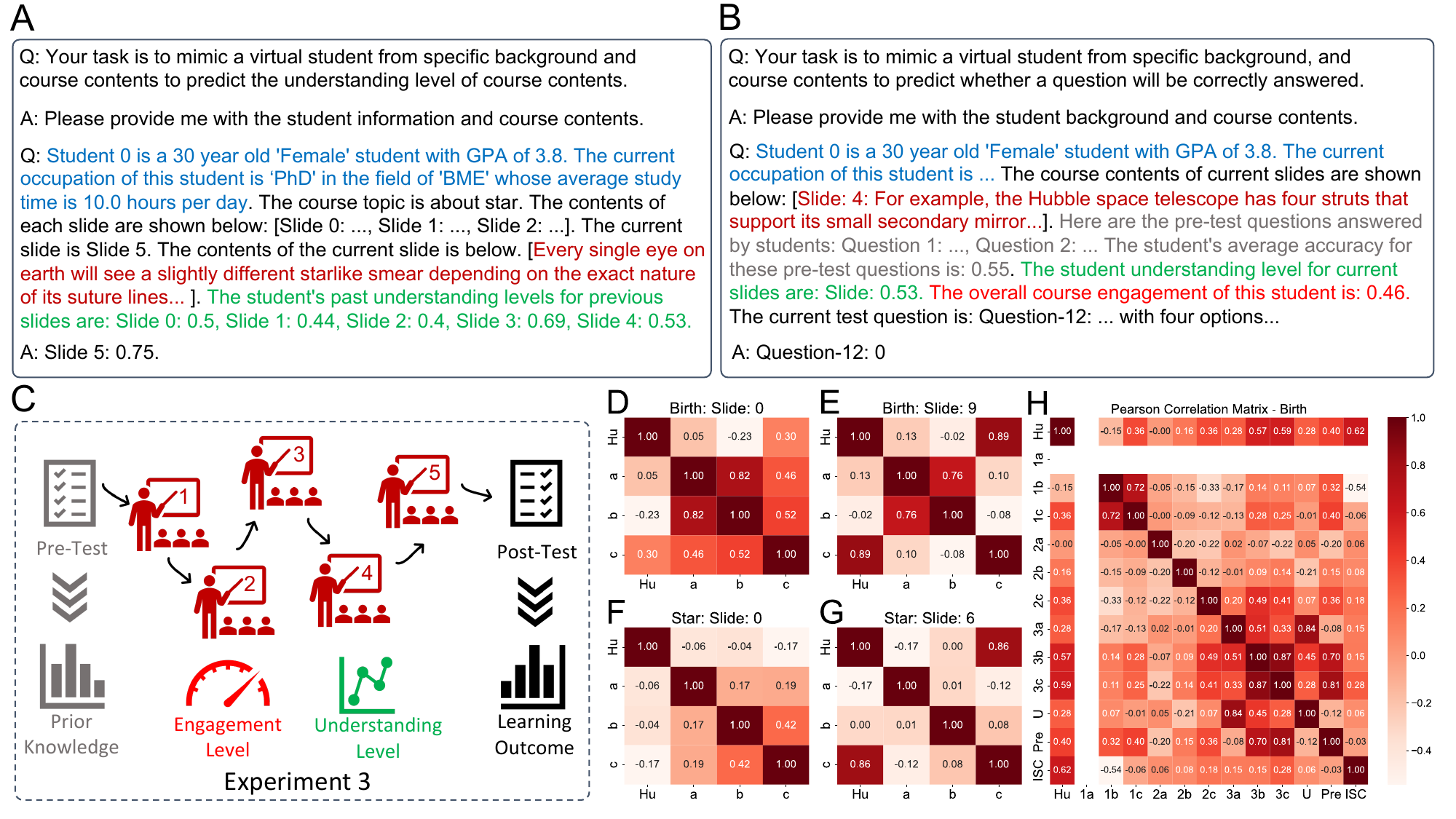}
\caption{(A)(B) depict an example of input prompts (denoted as Q) and LLMs' responses (denoted as A) in experiment 3: task 1 and task 2, respectively. Contents have been truncated. Full examples can be found in Appendix. (C). Illustration of information used for student simulation in experiment 3. (D)(E) depict Pearson correlation matrix in the first and last slide in $birth$ course in task 1 for student understanding level simulation, respectively. Hu represents real students' understanding level and a,b,c represent simulated understanding levels with Type a, Type b, Type c, respectively. (F)(G) depict Pearson correlation matrix in the first and last slide in $star$ course in task 1 for student understanding level simulation, respectively. (H) depict Pearson correlation matrix in $birth$ course in task 2 for student post test performance simulation. Hu represents real students' post test scores and 1a to 3c represent simulated post test scores with Type 1a to Type 3c, respectively. U, Pre, and ISC represent real students' average understanding levels, pre-test scores, and course engagement, respectively.
}
\label{f3}
\end{figure*}

The second experiment aims to go into a finer-grained level by simulating students with both demographics and past assessment performance(Fig.\ref{f2}(B)). We use an N = 4524 dataset of students in UK with more than five assessments from Open University Learning Analytics Dataset \cite{kuzilek2017open}. Fig. \ref{f2}(G) depicts the student number distribution in each region of UK. The final exam score is a numeric value from 0 to 100. Each student is simulated with 8 configurations, resulting in more than 40,716 responses of LLMs. We use temperature parameter to be 0 in this experiment for more deterministic outputs. The example of prompts to instantiate virtual students and predict final exam score according to both demographic information and assessment history is described in Fig. \ref{f2}(A). 
In order to investigate the impact of different factors on student simulation performance, we use different configurations to instantiate virtual students: Type 1: demographics, Type 2: all past assessments scores, Type 3: demographics + all past assessments scores, Type 4 to Type 8 refers to the configuration using only past one to past five assessments scores without demographics (Fig. \ref{f3}). 

We find highly aligned correlation in final exam scores between virtual and real students using Type ii ($r$ $=$ 0.69) and Type iii ($r$ $=$ 0.68), as depicted in Fig. \ref{f2}(C). More importantly, we find the correlation becomes stronger when we use more assessments scores as input (from Type iv to Type viii) (Fig. \ref{f2}(C) and appendix Fig), suggesting that students simulation tends to be more realistic as taking more student assessment history data for instantiating virtual students. 

Surprisingly, we find that the simulation performance gets a little worse when we add demographics (Type iii: $r$ $=$ 0.68) compared with only using assessments (Type ii: $r$ $=$ 0.69). The reason is mainly because of limited demographics in this dataset and some demographics may not directly reflect learning performance, which is different from experiment 1 that has rich demographics. This is supported by the inconsistent mapping results between IMD band \footnote{IMD (Indices of Multiple Deprivation) is a set of indices used in the UK to measure multiple dimensions of deprivation, including income, employment, health, education, crime, etc \cite{IMDcite}.} (Fig. \ref{f2}(H)) and students' final exam scores (Fig. \ref{f2}(J)).

However, we find consistency between students' average assessments (Fig. \ref{f2}(I)) and final scores (Fig. \ref{f2}(J)), which suggests and explains the capability of leveraging assessments for final exam score simulation. As a result, we observe highly aligned final scores between virtual students powered by Type iii (demographics + assessments) (Fig. \ref{f2}(K)) and real students (Fig. \ref{f2}(J)) in the distribution of regions in UK. However, if we only use demographics to instantiate students, the simulation performance drops heavily, as depicted in Fig. \ref{f2}(D)(E)(F), echoing Type i results in Fig. \ref{f2}(C) and suggesting importance of assessments information.

Although the map results look quite similar between average assessments (Fig. \ref{f2}(I)) and simulation results (Fig. \ref{f2}(K)), the range of color bar is different. More apparent difference could be found in Fig. \ref{f2}(D)(E)(F), suggesting that LLMs could capture correlation between assessments and final  scores but will not simply copying results from assessments for final score simulation. From these figures, we also observe that Type iii simulation exhibits consistent but not overlapped distribution among regions compared with real students. 
This diversity and divergence in simulation outcomes could contribute to the development of a more effective digital twin, as determinism alone may not adequately represent the range of student learning behaviors.

\begin{figure*}
\centering
\includegraphics[width=1\linewidth]{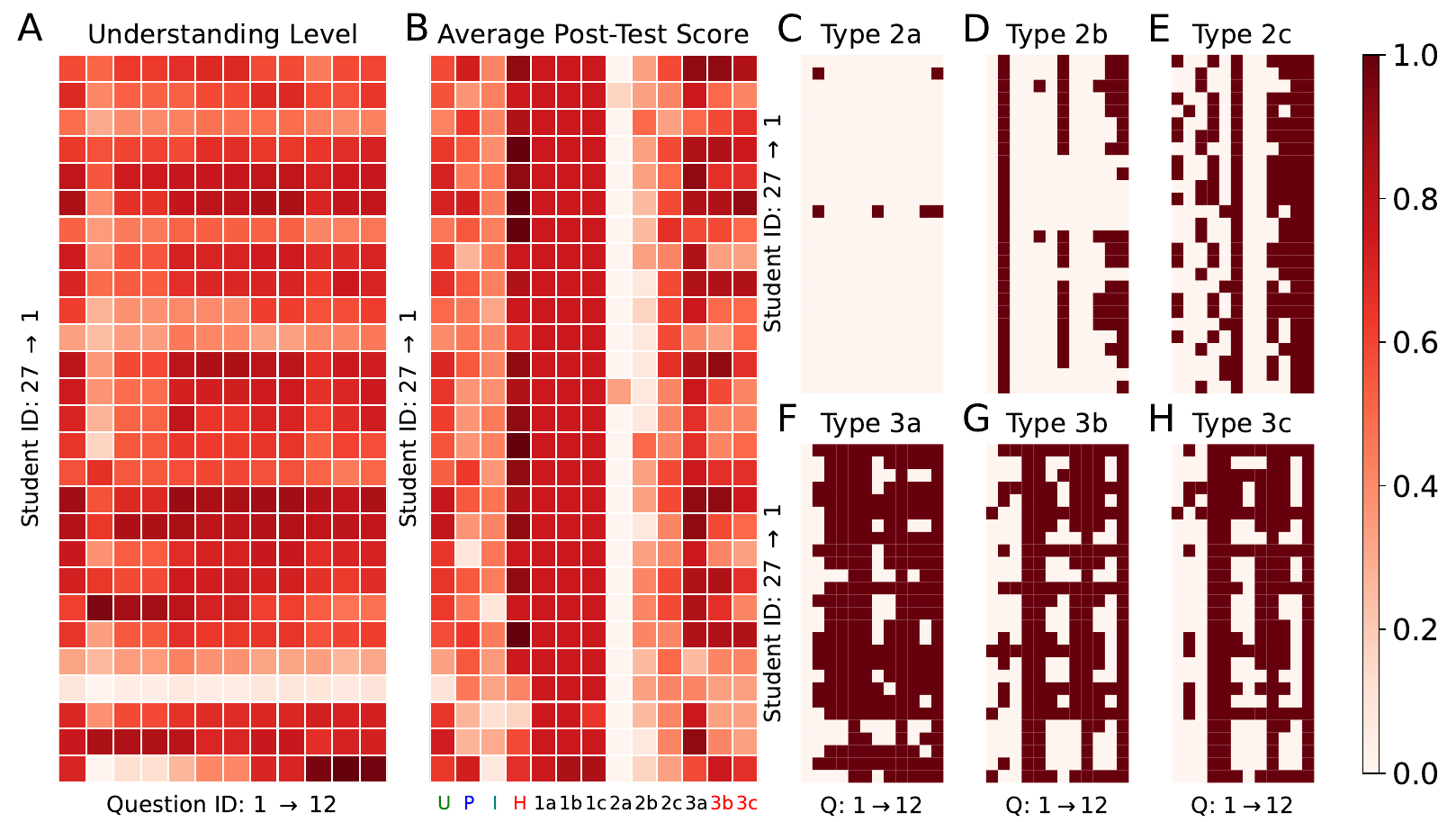}
\caption{(A). Each cell represents a specific real student's (vertical axis) understanding level (color depth) of specific slides related to the specific post-test question (horizontal axis). (B). Each cell represents a specific student. H represents real students' post test scores and 1a to 3c represent simulated post test scores with Type 1a to Type 3c, respectively. U, P, and I represent real students' average understanding levels, pre-test scores, and course engagement, respectively. (C)(D)(E)(F)(G)(H) represent each simulated student's answer correctness in each post-test question (horizontal axis) using Type 2a, 2b, 2c, 3a, 3b, 3c, respectively. Darker cell means virtual students answer this question correctly and vice versa.
All matrices share the same color bar.
}
\label{f4}
\end{figure*}

\subsection*{Experiment 3}

The third experiment aims to go into a much finer-grained level by simulating students with not only demographic information but also each student's interactions with course materials. We use a dataset of N = 27 students from \cite{madsen2021synchronized} which collected students' gaze data, pre-test and post-test scores while students were watching course videos. 
The dataset\cite{madsen2021synchronized} has suggested the synchronized eye movements among engaged students in the course and demonstrated the correlation between students' intersubject correlation (ISC) of gaze behaviors and post-test performance. Therefore, we use ISC to represent the engagement of students in our simulation. In the dataset, we also find a negative correlation between pupil size and student performance in a preliminary analysis. In addition, larger pupil size has been shown to be correlated with students' increased stress and overwhelming cognitive load \cite{chen2014using}. Therefore, we use pupil size to represent students' real-time understanding level during video watching in the course and larger pupil size will represent a lower understanding level in course concepts.

There are two courses named \textbf{birth} and \textbf{star}, which are segmented into 10 and 6 slides respectively according to course materials. Virtual students will not only simulate post-test scores as learning outcome, but also understanding levels after each slide in the course video. We use temperature parameter to be 0 in LLMs for more deterministic outputs.

\subsubsection*{Task 1: understanding level simulation}

We ask LLMs to mimic students with specific demographics and past understanding levels in previous slides, and then predict the understanding level in the current slide, which is a numeric value from 0 to 1. One example prompt and LLMs' responses are described in Fig. \ref{f3}(A). We use different configurations to instantiate virtual students to investigate the impact of different factors: Type a  simply inputs demographics and all course materials to predict understanding level of all slides together. Type b emphasizes the current slide material in addition to Type a's input, and LLMs will just predict understanding level of the individual slide. Type c will add all past understanding levels before the current slide in addition to Type b's input to predict understanding level of the individual slide. 

We find higher correlation in understanding levels between virtual and real students using Type c compared with Type a and Type b (Fig. \ref{f2}), suggesting that past understanding levels (instead of demographics) will significantly affect the understanding level in the current slide. Additionally, we observe increasingly improved simulation performance with the increase of slide number (Appendix Figure). For example, in Fig. \ref{f2}, the correlation between virtual (Type c) and real students increases significantly from the first slide to the last slide in both course \textbf{birth} ($r$ $=$ 0.30 to $r$ $=$ 0.89) and course \textbf{star} ($r$ $=$ -0.17 to $r$ $=$ 0.86). This suggests that more past understanding levels could provide more in-context information to help LLMs mimic more realistic students learning behaviors.

\subsubsection*{Task 2: learning outcome simulation}

We use prompts to instantiate virtual students and simulate post-test scores as learning outcomes according to demographics, understanding levels per slide, pre-test score (prior knowledge background), and engagement. One example is depicted in Fig. \ref{f3}(B). Similarly, we use different configurations to instantiate virtual students: Type 1a, 1b, 1c ask LLMs to predict the average accuracy in all post-test questions. Type 1a inputs demographics, all slides, and post-test questions. Type 1b adds past understanding levels of all slides in addition to Type 1a's input. Type 1c adds pre-test questions and average score in addition to Type 1b's input. The input of Type 2a, 2b, 2c follows the design of Type 1a, 1b, 1c, respectively. But Type 2a, 2b, 2c ask LLMs to predict accuracy of each question in post-test. However, note that Type 1a to 1c still input all post-test questions together. By contrast, Type 3a, 3b, 3c go into a much finer-grained level where LLMs will treat each question individually for single input and predict accuracy of each specific question. Type 3a inputs demographics, slides related to the specific post-test question, understanding levels of related slides, and the current post-test question. Type 3b adds pre-test questions and average score compared with Type 3a's input. Type 3c adds course engagement in addition to Type 3b's input. More details are depicted in appendix.

We find best aligned correlation in Type 3c between virtual and real students in course \textbf{birth}: $r$ $=$ 0.59 and \textbf{star}: $r$ $=$ 0.47. Fig. \ref{f3}(H) and appendix depicts the Pearson correlation matrix in \textbf{birth} and \textbf{star} respectively. Specifically, introducing pre-test information will significantly improve simulation performance from Type 1b ($r = -0.15$) to Type 1c ($r = 0.36$), Type 2b ($r = 0.16$) to Type 2c ($r = 0.36$), and Type 3a ($r = 0.28$) to Type 3b ($r = 0.57$). This is evidenced by correlation ($r = 0.40$) between pre-test and post-test performance. Additionally, we observe importance of understanding level and course engagement for more realistic simulation, from Type 2a ($r = 0$) to Type 2b ($r = 0.16$) and from Type 3b ($r = 0.57$) to Type 3c ($r = 0.59$), evidenced by correlation between understanding level and post-test performance ($r = 0.28$) and correlation between engagement and post-test performance ($r = 0.62$). A more intuitive visualization of simulation performance across all types for each student is depicted in Fig. \ref{f4}(B). These findings suggest the importance of incorporating students' previous knowledge background, real-time concept understanding level and course engagement for more realistic simulation.

More importantly, we observe that simulation gets more realistic when we treat each question individually (Type 3a to 3c: best $r$ $=$ 0.59) compared with treating all questions as a whole (Type 1a to 2c: best $r$ $=$ 0.36). To reveal such individual differences, we visualize virtual students' answers (correct or not) for each question in each cell with different types(Fig. \ref{f4}(C,D,E,F,G,H)). We find that simulation just fails if we only input demographics and course materials (Type 2a) since almost all questions are predicted to be wrongly answered. This is due to the requirement of more in-context information during course interactions. This explanation is supported by the improvement of simulation when understanding levels are introduced (Type 2b). Finally, the introduction of pre-test information and course engagement also further improves the variety of student simulation in Type 2c, 3b, 3c (Fig. \ref{f4}), contributing to developing a more effective digital twin.

\section*{Discussion}
Education is fundamental to the success of humans and society \cite{felten2020relationship}. Student simulation stands as the key for adaptive learning experience tailoring to meet unique needs of diverse students\cite{truong2016integrating}. Inspired by success of LLMs in cognitive studies \cite{binz2023using}, we leverage LLMs to mimic students with specific demographics and fine-grained interactions with course materials, underscoring the potential for digital twins to revolutionize the educational landscape in the future. 

Specifically, we observe that LLMs could capture the correlation between students' demographics and their potential success in the future exam. In addition, past assessments history could provide more helpful insights to support more realistic learning behaviors simulation. Furthermore, finer-grained simulation in the interaction with course materials could better predict the success of students in the future post-tests. Since we do not use training data to fine-tune LLMs, one question is why LLMs could work so well in such scenarios. One answer stems from LLMs' robust in-context learning abilities \cite{dong2022survey} and extensive knowledge base with human feedback \cite{lampinen2022can,stiennon2020learning}, making it understand the potential correlation between various factors and students' experience and success.

We highlight our unique contributions in two significant ways: First, we pioneer the validation of leveraging generative AI to simulate student learning behaviors in large-scale datasets and fine-grained course interaction experience. Second, rather than merely testing whether LLMs could solve a task or not, we emphasize exploring and explaining how LLMs could work well in student simulation via carefully designed probes and ablation study configurations in micro benchmarks, revealing versatile capabilities of LLMs across different contexts. As a result, our research significantly complements existing work.


We acknowledge potential limitations and open questions. First, the simulation is based on general understanding of LLMs so the accuracy may vary for individual cases, although this might not pose issues in practical scenarios like new teacher training. Second, LLMs' sensitivity to prompt design \cite{zhao2021calibrate} could lead to varied results when information order changes. Lastly, token size constraints \cite{jozefowicz2016exploring} might limit applications in extended educational contexts. We anticipate that future developments in LLMs may address these challenges.


In conclusion, we demonstrate the feasibility of leveraging generative AI to create digital twins for simulating student learning behaviors. Our research reveals strong alignment between virtual and real students' behaviors. By enabling the simulation and study of learning behaviors, this work has the potential to enhance learning experiences and outcomes, leading to transformative impacts on education and training. These findings offer educators deeper insights into dynamics of the learning environment, driving innovation in pedagogical strategies and inspiring the development of adaptive and personalized educational technologies. As such, our research contributes to the evolving landscape of online education.

\section*{Methods}

More details are depicted in Appendix.

\subsection*{Data and Code Availability}

All data and codes to reproduce the findings are available at: https://github.com/songlinxu/EduTwin.

\bibliography{sample}

\end{document}